%% file: iclr2023_conference.tex
\documentclass{article} 
\usepackage{iclr2023_conference,times} 

\input{math_commands.tex}

\usepackage{hyperref}
\usepackage{url}
\usepackage{graphicx}
 \usepackage{multirow}
 \usepackage{array}
 \usepackage{subfig}
\newcolumntype{C}[1]{>{\centering\arraybackslash}p{#1}}

\title{Multi-Modal Perceiver Language Model for Outcome Prediction in Emergency Department }

\author{Sabri Boughorbel \\
Qatar Computing Research Institute\\
Hamad Bin Khalifa University, Qatar\\
\AND
Fethi Jarray\\
LIMTIC Laboratory, UTM University, Tunis, Tunisia  \\
Higher institute of computer science of Medenine, Gabes University,  Tunisia\\
 \AND
Abdulaziz Al Homaid \\
Qatar Computing Research Institute\\
Hamad Bin Khalifa University, Qatar\\
\AND
Rashid Niaz \\ 
Medical Informatics Department \\
Sidra Medicine, Qatar \\
\AND
Khalid Alyafei \\ 
Medical Informatics Department \\
Sidra Medicine, Qatar 
 }

 \iclrfinalcopy

\begin{document}

\maketitle

\begin{abstract}
Language modeling have shown impressive progress in generating compelling text with good accuracy and high semantic coherence. An interesting research direction is to augment these powerful models for specific applications using contextual information. In this work, we explore multi-modal language modeling for healthcare applications. We are interested in outcome prediction and patient triage in hospital emergency department based on text information in chief complaints and vital signs recorded at triage.  We adapt Perceiver - a modality-agnostic transformer-based model that has shown promising results in several applications. Since vital-sign modality is represented in tabular format, we modified Perceiver position encoding to ensure permutation invariance. We evaluated the multi-modal language model for the task of diagnosis code prediction using MIMIC-IV ED dataset on 120K visits. In the experimental analysis, we show that mutli-modality improves the prediction performance compared with models trained solely on text or vital signs. We identified disease categories for which multi-modality leads to performance improvement and show that for these categories, vital signs have added predictive power. By analyzing the cross-attention layer, we show how multi-modality contributes to model predictions. This work gives interesting insights on the development of multi-modal language models for healthcare applications.
\end{abstract}
\section{Introduction}
 Large language models have recently made significant progress towards general-purpose applications of artificial intelligence \cite{ouyang2022training, adiwardana2020towards}. A combination of model architecture, large collections of text datasets and well-designed recipes of model training have been crucial elements for the fast progress in this field. An important active research is the extension of language models with other modalities. Recently, impressive results on image generation from text have been achieved \cite{ramesh2022hierarchical}. The combination of language, image, video and audio information is also a hot research area of multi-modal language models \cite{baevski2022data2vec}. In healthcare, the Electronic Health Records (EHR) capture multi-modal and complementary patient information \cite{acosta2022multimodal}. EHRs collect free-text clinical notes, numerical lab results  and other types of modalities such as medical imaging, structured diagnosis codes, wearable data or genomic diagnostic information. The application of multimodal approaches to healthcare has a high potential to speed-up the adoption of AI tools in clinical practice. In the following paragraph, we provide a brief overview of multi-modal approaches and justify the chosen direction for this work.

Multimodal approaches can be categorized into  early (input), intermediate (feature) and late fusion (decision) \cite{ramachandram2017deep, xu2022multimodal}. Early fusion combines the input data into a single representation before being fed into deep networks. The main drawback of these techniques is that the subsequent classification layers can not differentiate between the different modalities.  In intermediate fusion, each modality is first processed separately to extract features, and then features are combined from different modalities into an intermediate representation.  It can be implemented  by various schemes such as  concatenation, summation and  attention mechanisms.   Baevski et al.  proposed a transformer-based framework, data2vec, for learning from speech, vision and language data \cite{baevski2022data2vec}. Every modality  is converted into a sequence of inputs. Image is encoder by ViT-strategy \cite{dosovitskiy2020image},  speech is encoded using a multi-layer 1-D CNN and text is tokenized by  byte-pair encoding. The model is trained in a self-supervised manner by following masking strategies similar to BERT \cite{devlin2018bert}.  Data2vec has the disadvantage of being computationally expensive.
Late fusion approaches separately process the  different modalities and then combine the outputs at the last layers. The final decision can be made by simple strategies such as majority voting or averaging \cite{deng2020multimodal}. This approach is useful when the different modalities are relatively independent and thus reduces the risk of information loss.

Perceiver has been proposed as a modality-agnostic transformer to address the computational bottleneck in self-attention \cite{jaegle2021perceiver}. As shown in Figure~\ref{fig:perceiver}, Perceiver introduces a latent vector of fixed size $N$. The modality fusion in Perceiver is hybrid. The input data is fused at different stages of the network. Hence the model has different opportunities, throughout its depth, to extract additional pieces of information from the various modalities. The cross-attention layer attends the input data of size $M$ with the latent array. This operation is computationally cheap and allows to decouple the model depth from the input data size. The goal of this work is to apply Perceiver for the task of predicting diagnosis code in emergency department (ED) visits using both text information from the chief complaint note and numeric information from patient state and vital signs. Early prediction of patient outcome in ED allows for a better planning of the clinical workforce and an improved management of care demand in ED.

\section{Methods}
\begin{figure}
   \centering
    \includegraphics[scale=0.5]{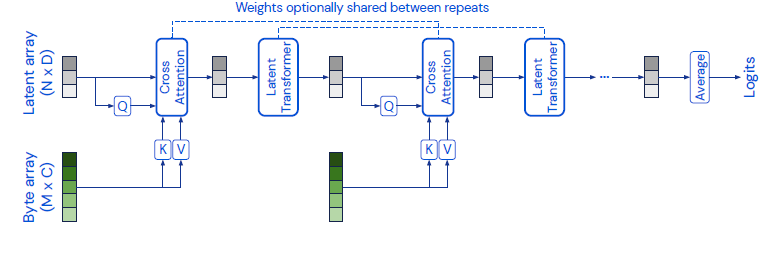} 
   \caption{Perceiver architecture. The input data in green is fused at different depths using the cross-attention block (figure from \cite{jaegle2021perceiver}).}
    \label{fig:perceiver}
\end{figure}

We use Perceiver as the core model for the integration of text data and vital signs for the classification task. The vital signs are represented as numeric arrays and text data are represented by a learned embeddings part of Perceiver training. For each visit, both representations are concatenated and fed to the different blocks of Perceiver as shown in Figure~\ref{fig:perceiver}. In the following, we describe the dataset we used and how multiple modalities are combined in Perceiver. Through different experiments and ablation study, we illustrate the role of two modalities (text, vital signs) in the prediction task. 



\subsection{Dataset}
We used MIMIC-IV ED dataset for our experiments \cite{johnson2021mimic}. We included visits with ICD-10 codes. As the triage information is very limited, we considered the prediction of a high-level ICD code without decimals. We focused on prediction primary diagnosis and discarded secondary ones. We limited our analysis to the top-50 most frequent codes in the dataset to have representative data for model training \cite{kaur2021systematic}.  The text data from the  prediction ICD codes in the Emergency Department using both chief complaints and vital signs recorded at admission. The chief complaints contains  patient’s reported reason for presenting to ED. It is in the form of a short free-text. The data is pre-processed  by removing punctuation, case lowering and tokenization. The tabular numeric data comprises 6 vital signs (temperature,	heart rate,	respiration rate,	oxygen saturation, 	systolic blood pressure and	dystolic blood pressure) as well as two variables assessed by nurses: pain level scored on a scale of 1 to 10 and acuity (severity) scored on a scale of 1 to 5 where 1 indicates the highest severity and 5 indicates the lowest.
\subsection{Model Setup and Training}
Since the text length is short in our application, we chose a compact embedding with a size of 16. Also, we wanted to maintain a balance between the representation size of the tabular data and the text data. The vital signs represented as 8-dimensional array is used without additional processing. Both representation from text and vital signs are concatenated and used as input to Perceiver model. The depth of the model is chosen to 4. Hence the cross-attention block and latent transformer are repeated four times. The size of latent array is chosen to be 16. The different modalities are  merged at each cross-attention block. In the experiments, we analyzed the cross-attention layers to investigate the role of the different modalities in the prediction. We split the dataset of 120K visits into training and test sets. Perceiver model is trained for 50 epochs until the cross-entropy loss has stabilized. All results presented in the tables are obtained on the test sets. Model training is repeated 5 times with random splits of training and test sets. The mean and standard deviations of the metrics across the  data randomization are reported in the tables.
 
\section{Results and discussion}
We investigated the role of position encoding in Perceiver for the tabular modality.  Table~\ref{table:results2} provides a comparison between Perceiver trained with and without position encoding for the tabular modality (vital signs). The latter ensures permutation  invariance and leads to improved results. Permutation-invariant transformers have shown good results for tabular data \cite{dash2022permutation}. 
\begin{table}[h]
\centering
\begin{tabular}{l|ccccc}
\textbf{Model}    & \textbf{Prec} & \textbf{Rec}  & \textbf{F1}   & \textbf{Acc}  & \textbf{AUC}  \\ \hline 
\textbf{Perceiver-WPE} & \bf{74$\pm$0.35} & \bf{50$\pm$0.28} & \bf{60$\pm$0.13} & \bf{43$\pm$0.13} & \bf{96$\pm$0.04} \\
\textbf{Perceiver-PE}     & 73$\pm$0.21 & 48$\pm$0.78 & 57$\pm$0.37 & 39$\pm$0.26 & 94$\pm$0.31 \\       
\end{tabular}
\caption{Comparison of Perceiver without position encoding (WPE) on the tabular modality (vital signs) and Perceiver with position encoding (PE) using Fourier features. The values are micro averaged.}
\label{table:results2}
\end{table}

In the subsequent, we adopt a Perceiver without position embedding for vital signs. Table~\ref{table:results} summarizes the results of predicting the primary diagnosis code using the different modalities. It shows that by combining the chief complaint notes and vital signs, the prediction performance are better for all metrics (Prec, Rec, F1, ACC, AUC) than using single text features  or vital signs.
\begin{table}[h]
\centering
\begin{tabular}{l|ccccc}
\hline
& \multicolumn{5}{c}{\textbf{Macro}}                                            \\ 
\textbf{Modality}    & \textbf{Prec} & \textbf{Rec}  & \textbf{F1}   & \textbf{Acc}  & \textbf{AUC}  \\ \hline
\textbf{Text+Vitals} & \bf{66$\pm$0.27} & \bf{38$\pm$0.36} & \bf{48$\pm$0.28} & \bf{32$\pm$0.34} & \bf{94$\pm$0.10}  \\
\textbf{Text}        & 60$\pm$1.17 & 29$\pm$0.84 & 39$\pm$0.97 & 25$\pm$0.76 & 92$\pm$0.43 \\
\textbf{Vitals}        & 39$\pm$1.72 & 16$\pm$1.67  & 21$\pm$1.54 & 12$\pm$1.30 & 80$\pm$0.28 \\ \hline
& \multicolumn{5}{c}{\textbf{Micro}}                                            \\ 
\textbf{Modality}            & \textbf{Prec} & \textbf{Rec}  & \textbf{F1}   & \textbf{Acc}  & \textbf{AUC}  \\ \hline
\textbf{Text+Vitals} & \bf{74$\pm$0.35} & \bf{50$\pm$0.28} & \bf{60$\pm$0.13} & \bf{43$\pm$0.13} & \bf{96$\pm$0.04} \\
\textbf{Text}        & 76$\pm$0.14 & 43$\pm$0.61 & 55$\pm$0.51 & 37$\pm$0.48 & 95$\pm$0.23 \\
\textbf{Vitals}        & 46$\pm$0.21 & 27$\pm$0.42 & 32$\pm$1.24  & 21$\pm$1.31 & 71$\pm$0.4 \\ \hline
\end{tabular}
\caption{Performance of single modalities vs. multi-modalities. The task is a multi-class classification of ICD codes. The metrics are Precision, Recall, F1 score, Accuracy and Area Under ROC Curve. The metrics are macro and micro averaged.}
\label{table:results}
\end{table}

\begin{table}[h]
\centering
\begin{tabular}{C{2cm}|C{1.5cm}C{1.2cm}lC{6cm}}
&   \textbf{AUC Diff.} & \textbf{ICD-10}     & \textbf{Description}                                       \\ \hline 
\multirow{5}{*}{\textbf{Top Codes}} & 5.6                           & A41                     & Sepsis \\
& 5.4                           & E87                     & Disorders of fluid and acid-base balance \\
& 4.0                           & R20                     & Disturbances of skin sensation                              \\
& 3.9                           & I48                     & Atrial fibrillation and flutter                             \\
& 3.1                           & I63                     & Cerebral infarction                                         \\ \hline
\multirow{5}{*}{\textbf{Bottom Codes}}                  & 0                             & D64                     & Other anemias                                               \\
& 0                             & J45                     & Asthma                                                      \\
& 0                             & T78                     & Adverse effects, not elsewhere classified                   \\
& 0                             & R45                     & Symptoms and signs involving emotional state                \\
& 0         & S72 & Fracture of femur     \\ \hline                                     
\end{tabular}
\caption{Performance difference (in terms  of Macro AUC) between  multi-modal Perceiver (text+vitals) and text-only model. The results are stratified by disease category. }

\label{table-comparison}
\end{table}

\begin{figure}[htbp]
 \centering
\subfloat[Average cross-attention matrix from the 4th (last) Perceiver block. T0-T7 denote token position.]{\includegraphics[scale=0.5]{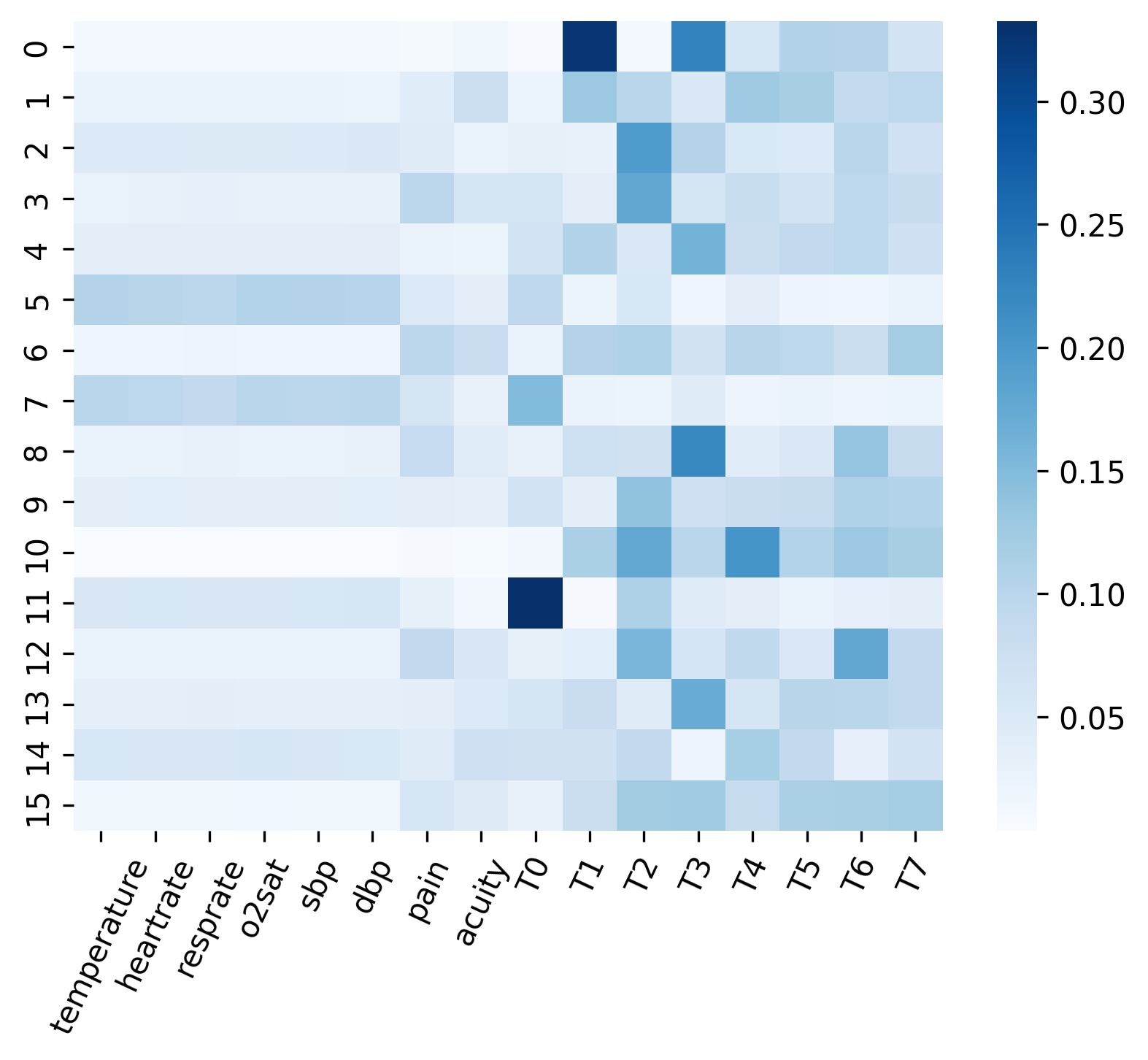}}
\subfloat[Cross-attention matrix of an ED visit data having an ICD Code R50 (Fever). The visit text note is 'fever, ili, right-sided abdominal pain'.  ILI means (Influeza Like Illness)]{\includegraphics[scale=0.5]{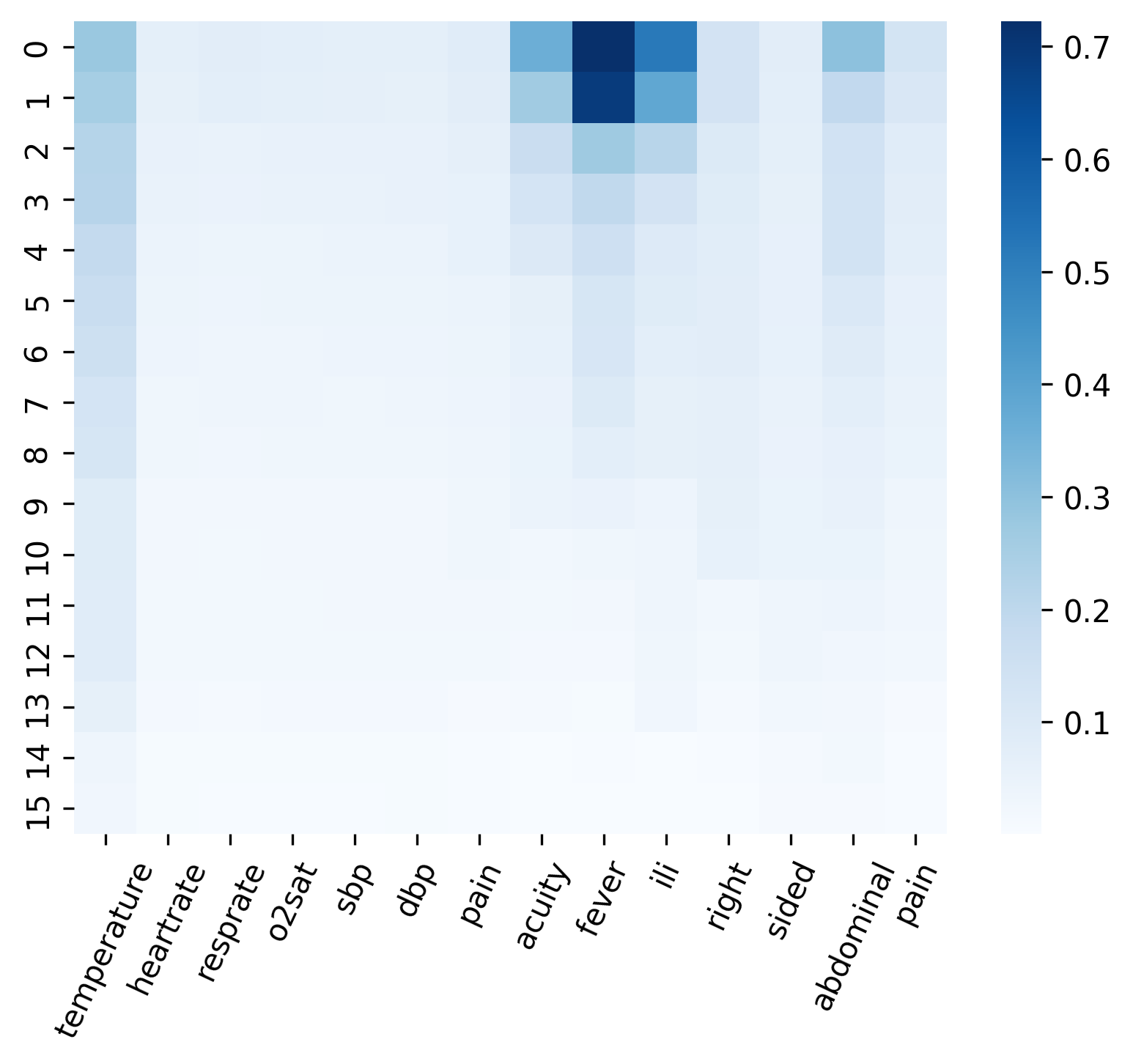}}
\caption{(a) From Vital signs, pain and acuity are the most contributors. The text modality has a strong cross attention with no clear pattern on token position. (b) The chief complaint of the visit with fever symptoms.}
\label{fig:crossattention}
\end{figure}

For a better understanding of  modality roles, we stratified the performance results by disease category (ICD codes). We present in Table~\ref{table-comparison}, the performance difference between multi-modal and single modal (text) models. The largest performance improvement was reported for disease such as sepsis, atrial fibrillation or cerebral infarction. This could be explained by the complementary information recorded in the vital signs and text for the diagnosis of these diseases. Previous work  quantified vital-signs symptoms as diagnostic features for these diseases \cite{di2010fluid, hunt2022sepsis, rienstra2012symptoms, greenlund2003low}. The bottom codes in Table~\ref{table-comparison} correspond to diseases such as anemia, asthma, fracture or adverse effects. These conditions, except asthma, are not typically diagnosed using the measured vital signs. For asthma, we expected that oxygen saturation should have a higher predictive contribution.  This could possibly be explained by the variability of symtomps and vital signs at admission. Figure~\ref{fig:crossattention} gives a visual representation of the cross-attention layer which is responsible of fusing the two modalities. Figure~\ref{fig:crossattention}-a shows that text modality is overall more activated than vital signs. Figure~\ref{fig:crossattention}-b reveals the contribution of temperature vital signs in Perceiver prediction for a Fever-related ED admission. 

\vspace*{-0.5cm}
\section{conclusion}
We investigated a multi-modal approach for learning from text and tabular data for the prediction of health outcome  in hospital emergency department. The method is based on Perceiever model which enables  early and intermediate fusion of  modalities.  The experimental results show the effectiveness of the multi-modal approach compared with single modalities. As a future work, we plan to expand the scale of the language models using larger datasets such as MIMIC-III ICU. We plan to explore the direction of fine tuning pre-trained  large language models  and extend them with additional modalities such as tabular or image data.

\bibliography{iclr2023_conference}
\bibliographystyle{iclr2023_conference}

\end{document}

%% file: math_commands.tex
\usepackage{amsmath,amsfonts,bm}

\def\eqref#1{equation~\ref{#1}}

\def\1{\bm{1}}

\DeclareMathAlphabet{\mathsfit}{\encodingdefault}{\sfdefault}{m}{sl}
\SetMathAlphabet{\mathsfit}{bold}{\encodingdefault}{\sfdefault}{bx}{n}

